%% file: main.tex
\documentclass[11pt,a4paper]{article}
\usepackage[hyperref]{emnlp2020}
\usepackage{times}
\usepackage{latexsym}

\usepackage{subcaption}
\usepackage{amssymb}
\usepackage{microtype}
\usepackage{color, colortbl} 
\usepackage{xcolor}
\usepackage{mdframed}

\aclfinalcopy 
\def\aclpaperid{***} 

\usepackage{url}
\usepackage{booktabs} 
\usepackage{xspace}
\usepackage[normalem]{ulem}
\usepackage{amsmath}
\usepackage{amssymb}
\usepackage{color, colortbl} 
\usepackage{enumitem}
\usepackage{mathtools}
\usepackage[colorinlistoftodos]{todonotes}
\usepackage{comment}

\definecolor{purple}{rgb}{0.5,0,1}
\definecolor{dcyan}{rgb}{0.2,0.6,0.5}
\definecolor{darkgreen}{rgb}{0,200,0}
\definecolor{light-gray}{gray}{0.95} 
\definecolor{darkgreen}{RGB}{0,140,0}
\definecolor{darkred}{RGB}{200,0,0}
\definecolor{lightgreen}{RGB}{231,255,219}
\definecolor{lightred}{RGB}{252,231,234}
\definecolor{lightyellow}{RGB}{250,253,191}
\definecolor{DarkRed}{RGB}{130,25,0}

\newcommand{\dataset}{\mathcal{D}}

\newcommand{\clusterssymb}{\includegraphics[scale=0.04,trim=0.2cm 0cm 3.5cm 0cm, clip=false]{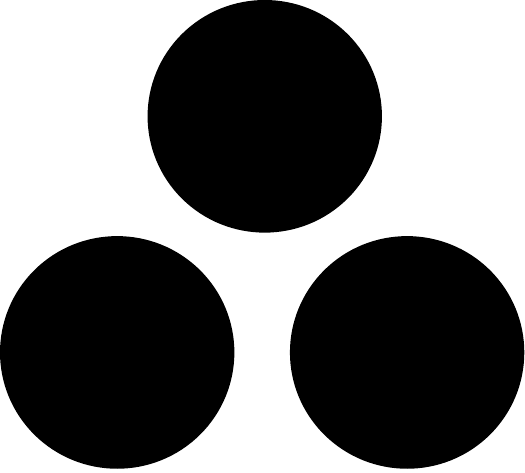} \;}

\newcommand{\boolq}{\textsc{BoolQ}\xspace}
\newcommand{\boolqpp}{\textsc{BoolQ++}\xspace}
\newcommand{\boolqR}{\textsc{BoolQ}{$_\text{\clusterssymb}$}\xspace}
\newcommand{\boolqc}[1]{\textsc{BoolQ}$^{#1}_\text{\clusterssymb}$\xspace}
\newcommand{\boolqExperts}{\textsc{BoolQ-$e$}{$_\text{\clusterssymb}$}\xspace}
\newcommand{\datasetSeed}{{\dataset}_S}
\newcommand{\datasetR}{\dataset{\color{darkgreen}_\text{\clusterssymb}}}
\newcommand{\namecite}[1]{\citeauthor{#1}~\shortcite{#1}}
\newcommand{\ignore}[1]{}
\definecolor{purple}{rgb}{0.5,0,1}
\definecolor{dcyan}{rgb}{0.2,0.6,0.5}
\definecolor{darkgreen}{rgb}{0,0.25,0}
\definecolor{light-gray}{gray}{0.95} 

\definecolor{DarkRed}{RGB}{130,25,0}
\newcommand{\daniel}[1]{#1}
\newcommand{\ashish}[1]{#1}
\newcommand{\tushar}[1]{#1}

\newcolumntype{L}[1]{>{\raggedright\let\newline\\\arraybackslash\hspace{0pt}}m{#1}}
\newcolumntype{C}[1]{>{\centering\let\newline\\\arraybackslash\hspace{0pt}}m{#1}}

\definecolor{purple}{rgb}{0.5,0,1}
\definecolor{dcyan}{rgb}{0.2,0.6,0.5}
\definecolor{light-gray}{gray}{0.95} 

\definecolor{darkgreen}{RGB}{0,140,0}
\definecolor{darkred}{RGB}{200,0,0}
\definecolor{lightgreen}{RGB}{218,237,213}
\definecolor{lightred}{RGB}{255,205,212}
\definecolor{lightyellow}{RGB}{255,240,160}
\definecolor{lightblue}{RGB}{195,221,255}

\newcommand{\roberta}{\textsc{RoBERTa}\xspace}

\newcommand{\multirc}{\textsc{MultiRC}\xspace}

\newcommand{\redtext}[1]{\colorbox{lightred}{#1}\xspace}
\newcommand{\greentext}[1]{\colorbox{lightgreen}{#1}\xspace}
\newcommand{\yellowtext}[1]{\colorbox{lightyellow}{#1}\xspace}
\newcommand{\bluetext}[1]{\colorbox{lightblue}{#1}\xspace}

\title{
    \vspace*{-0.5in}
    {{\small \hfill EMNLP'20}\\
    \vspace*{.25in}} 
    \emph{More Bang for Your Buck:} \\ 
    Natural Perturbation for Robust Question Answering
}

\author{
    Daniel Khashabi \and Tushar Khot \and Ashish Sabharwal \\
    Allen Institute for AI, Seattle, WA, U.S.A.  \\
 {\tt  \footnotesize
 \{danielk,tushark,ashishs\}@allenai.org 
 }  
}

\begin{document}

\maketitle

\begin{abstract}
Deep learning models for linguistic tasks require large training datasets, which are expensive to create. As an alternative to the traditional approach of creating new instances by repeating the process of creating one instance, we propose doing so by first collecting a set of seed examples and then applying human-driven \emph{natural perturbations} (as opposed to rule-based machine perturbations), which often change the gold label as well. Such perturbations have the advantage of being relatively easier (and hence cheaper) to create than writing out completely new examples. Further, they help address the issue that even models achieving human-level scores on NLP datasets are known to be considerably sensitive to small changes in input. To evaluate the idea, we consider a recent question-answering dataset (\boolq) and study our approach as a function of the \emph{perturbation cost ratio}, the relative cost of perturbing an existing question vs.\ creating a new one from scratch. We find that when natural perturbations are moderately cheaper to create (cost ratio under 60\%), it is more effective to use them for training \boolq models: such models exhibit 9\% higher robustness and 4.5\% stronger generalization, while retaining performance on the original \boolq dataset.


\end{abstract}

\section{Introduction}


Creating large datasets to train NLP models has become increasingly expensive.
While many datasets~\cite{bowman2015snli,Rajpurkar2016SQuAD10} targeting different linguistic tasks
have been proposed, nearly all are created by repeating a fixed process used for writing a single example. 
This approach results in many \emph{independent} examples, each generated from scratch. We propose an alternative, often substantially cheaper training set construction method where, after collecting a few seed examples, the set is expanded by applying human-authored \emph{minimal perturbations} to the seeds.

\begin{figure}[t]
    \centering
    \includegraphics[scale=0.44,trim=0.0cm 0.0cm 0cm 0.7cm, clip=false]{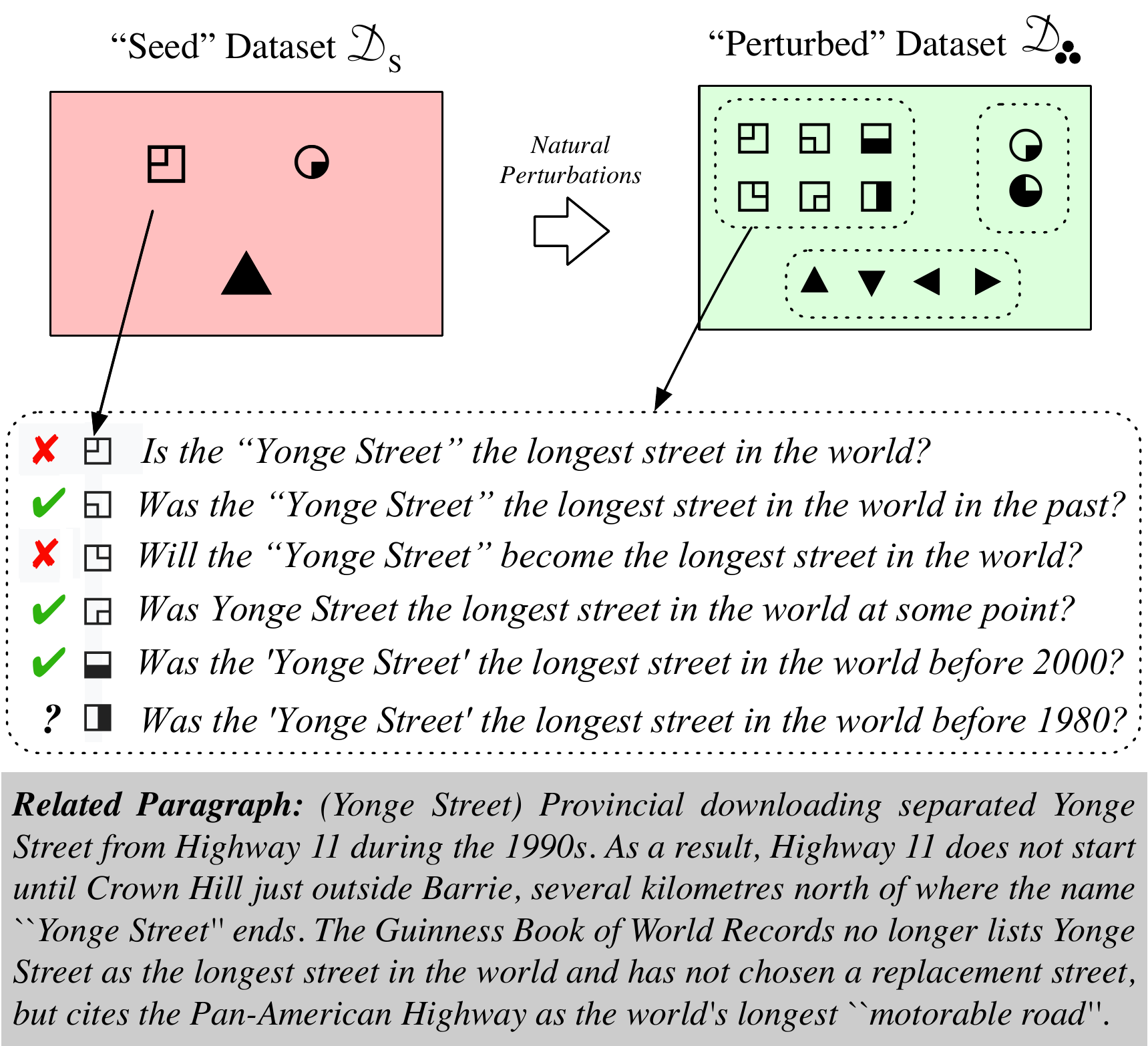}
    \caption{
    Training set creation via minimal-perturbation clusters. Left: Seed dataset $\datasetSeed$ with 3 instances (shown as different shapes). Right: Expanded dataset $\datasetR$ with 10 instances, comprising 2-4 minimal-perturbations (illustrated as rotation, fills, etc.) of each seed instance. Human-authored perturbations aren't required to always preserve the answer (yes/no in the example) and often add richness by altering the answer.
    }
    \label{fig:example:intro}
\end{figure}

Fig.~\ref{fig:example:intro} illustrates our proposal of using \emph{natural perturbations}. We use the traditional approach to first create a small scale \emph{seed dataset} $\datasetSeed$, shown as the red rectangle on the left with three instances (denoted by different shapes). However, rather than simply repeating this process to scale up $\datasetSeed$ to a larger dataset $\dataset$, we set up a different task: ask crowdworkers to create multiple \emph{minimal perturbations} of each seed instance (shown as rotation, fills, etc.) with an encouragement to change the answer.
The end result is a larger dataset $\datasetR$ of a similar size as $\dataset$ but with an inherent structure: \emph{clusters} of \emph{minimally-perturbed} instances with mixed 
labels, denoted by the green rectangle at the right in Fig.~\ref{fig:example:intro}.

An inspiration for our approach is the lack of robustness of current state-of-the-art models to minor adversarial changes in the input~\cite{Jia2017AdversarialEF}. We observed a similar phenomenon even with model-agnostic, human-authored changes to yes/no questions (as shown in Fig.\ref{fig:example:intro}), despite models achieving near-human performance on this task. Specifically, we found the accuracy of a \roberta model trained on \boolq~\cite{clark2019boolq} to drop by 15\% when evaluated on locally perturbed questions. These new questions were, however, no harder for humans.
This raises the question: \emph{Can a different way of constructing training sets help alleviate this issue?} Minimal perturbations, as we show, provide an affirmative answer.



Perturbing a given example is generally a simpler task, costing only a fraction of the cost of creating a new example from scratch. We call this fraction the \emph{perturbation cost ratio} (henceforth referred to as \emph{cost ratio}), and assess the value of our perturbed training datasets as a function of it. As this ratio decreases (i.e., perturbations become cheaper), one, of course, obtains a larger dataset than the traditional method, at the same cost. More importantly, even when the ratio is only moderately low (at 0.6), models trained on our perturbed datasets exhibit desirable advantages: They are 9\% more robust to minor changes and generalize 4.5\% better across datasets than models trained on \boolq.

Specifically, our \emph{generalization experiment} with the \multirc~\cite{khashabi2018looking} dataset demonstrates that models trained on perturbed data outperform those trained on traditional data when evaluated on unseen, unperturbed questions from a different domain. Second, we \emph{assess robustness} by evaluating on \boolqExperts~\cite{gardner2020contrastsets}, a test set of expert-generated perturbations that deviate from the patterns common in large-scale crowdsourced perturbations. Our zero-shot results here indicate that models trained on perturbed questions go beyond simply learning to memorize particular patterns in the training data. Third, 
\daniel{we find that training on the perturbed data, for the most part, }
continues to  \emph{retain performance} on the original task.

Even with the worst case cost ratio of 1.0 (when perturbing existing questions is no cheaper than writing new ones), models trained on perturbed examples remain competitive on all our evaluation sets. This is an important use case for situations that simply do not allow for sufficiently many distinct training examples (e.g., low resource settings, limited amounts of real user data, etc.). Our results at ratio 1.0 suggest that simply applying minimal perturbations to the limited number of real examples available in these situations can be just as effective as (hypothetically) having access to large amounts of real data.

In summary, we propose a novel method to construct datasets that combines traditional independent example collection approach with minimal natural perturbations. We show that for many reasonable cases, using perturbation clusters for training can result in cost-efficiently trained high-quality, robust models that generalize across datasets.

\section{Related Work}
\label{sec:related:work}

\paragraph{Data augmentation.}
There is a handful of work that studies semi-automatic contextual augmentation~\cite{kobayashi2018contextual,cheng2018towards}, often with the goal of creating better systems. We, however, study natural human-authored perturbations as an alternative dataset construction method. 
A related recent work is by \citet{kaushik2019learning}, who, unlike the goal here, study the value of natural-perturbations in reducing artifacts.

\paragraph{Adversarial perturbations.} 
A closely relevant line of work is \emph{adversarial perturbations} to expose the weaknesses of 
systems upon local changes and criticize their lack robustness~\cite{ebrahimi2018adversarial,glockner2018breaking,dinan2019build}. 
For instance, \namecite{tableilp2016:ijcai} showed significant drops upon perturbing answer-options for multiple-choice question-answering. 
Such rule-based perturbations have simple definitions leading to them being easily reverse-engineered by models~\cite{Jia2017AdversarialEF} and generally use label-preserving, shallow perturbations~\cite{rewriting}. In contrast, our natural human-authored perturbations are harder for models.\footnote{We tried the system by \citet{rewriting} on our questions, but it resulted in very limited variations. See Appendix \ref{supp:manualperturbations}.}
\daniel{
More broadly, adversarial perturbations research \ashish{seeks examples that stumble existing models, while our focus is on expanding datasets} in a cost-efficient way.
}

\paragraph{Minimal-pairs in NLP.}
Datasets with minimal-pair instances are relatively well-established in certain tasks, such as \emph{Winograd schema} datasets~\cite{levesque2011winograd,PengKhRo15,sakaguchi2019winogrande}, or the recent \emph{contrast sets}~\cite{gardner2020contrastsets}. 
However, the importance of datasets with \emph{pairs} (i.e., \emph{clusters} of size two) is not well-understood. 
Our findings about perturbation clusters could potentially be useful for the future construction of datasets for such tasks. 

\section{Expansion via Perturbation Clusters}
\label{sec:expansion}

Our approach mainly differs from traditional approaches in how we expand the dataset given seed examples. Rather than repeating the process to generate more examples, we apply minimal alterations to the seed examples, in two high-level steps:

\begin{figure}[ht]
    \centering
    \includegraphics[scale=0.49,trim=1cm 0.6cm 1.4cm 0.5cm, clip=false]{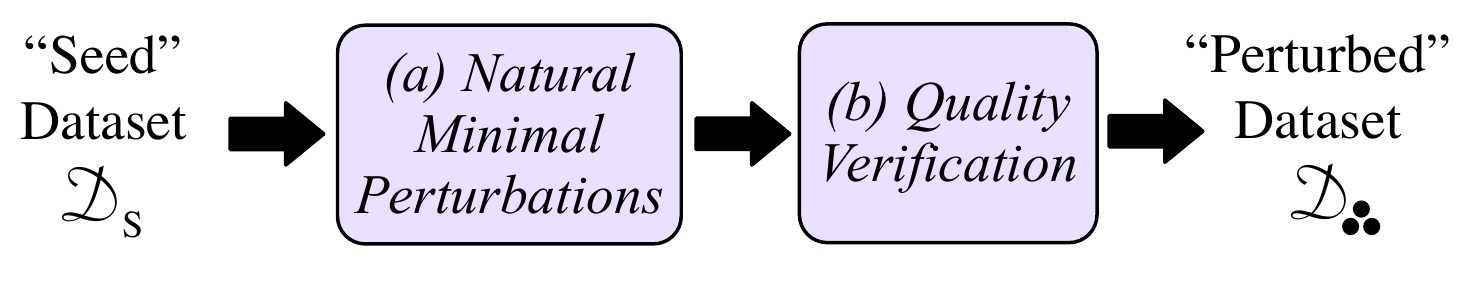}
\end{figure}

The first step generates the initial set of examples with natural perturbations. It should respect certain principles: (a) The construction should apply minimal changes (similar to the ones in Fig.~\ref{fig:example:intro}), otherwise the resulting clusters might be too heterogeneous and less meaningful. (b) A substantial proportion of natural perturbations should \emph{change} the answer to the questions. (c) It should incentivize creativity and diversity in local perturbations by, for instance, showing thought-provoking suggestions, using a diverse pool of annotators~\cite{geva2019we}, etc. The second independent verification step ensures dataset quality by (a) getting the true gold label and (b) ensuring all generated questions are answerable given the relevant paragraph, in isolation from the original question.

\paragraph{\boolqR: \boolq\ Expansion.}
We obtain $\datasetSeed$ by sampling  questions from \boolq~\cite{clark2019boolq}, which is a QA dataset where each boolean (``yes''/``no'' answer) question could be inferred from an associated passage. 
We then follow the above two-step process, resulting in \boolqR, a naturally perturbed dataset with $17k$ questions derived from $4k$ seed questions: 

\underline{\emph{a) minimal perturbations:}}
Crowdworkers are given a question and its corresponding gold answer based on supporting paragraph. 
Then the workers are asked to change the question in order to \emph{flip the answer to the question}. 
While making changes, the workers are guided to keep their changes minimal (adding or removing up to 4 terms) while resulting in proper English questions.  
Additionally, for each seed question, crowd-workers are asked to generate perturbations till the modified question is challenging for a machine solver (i.e., \roberta trained on \boolq, should have low confidence on the correct answer). 
Note that we do not require the model to answer the question incorrectly and not all questions are challenging for the model. Our main goal here is to encourage interesting questions by using the trained model as the guide. 


\underline{\emph{b) question verification.}}
Given the perturbed questions, we asked multiple annotators to answer these questions. These annotations served to eliminate ambiguous questions as well as those that cannot be answered from the provided paragraph. 
The annotation was done in two steps: (i) in the first step, we ask 3 workers to answer each question with one of the three options (``yes``, ``no`` and ``cannot be inferred from the paragraph''). We filtered out the subset of the questions that were not agreed upon (i.e., not a consistent majority label) or were marked as ``cannot be inferred from the paragraph'' by majority of the annotators.
To speed up the annotations, the annotation were done on a cluster-level, i.e., annotators could see all the different modified questions corresponding to a paragraph. (ii) subsequently, each modified questions is also annotated  \emph{individually} to ensure that questions can be answered in isolation
(as opposed to answering them while seeing all the questions in a cluster.) 
The annotations in this step only have two labels (``yes''/``no'') and again questions that were not agreed upon were filtered. 

Sample questions generated by our process are shown in Fig.~\ref{fig:example:intro}. 
We evaluate the impact of perturbations via this dataset. 

\begin{figure*}[tb]
    \centering
    \begin{subfigure}[b]{0.32\textwidth}
        \centering
        \caption{\boolqExperts}
        \includegraphics[scale=0.48,trim=0.2cm 0.5cm 0cm 0cm]{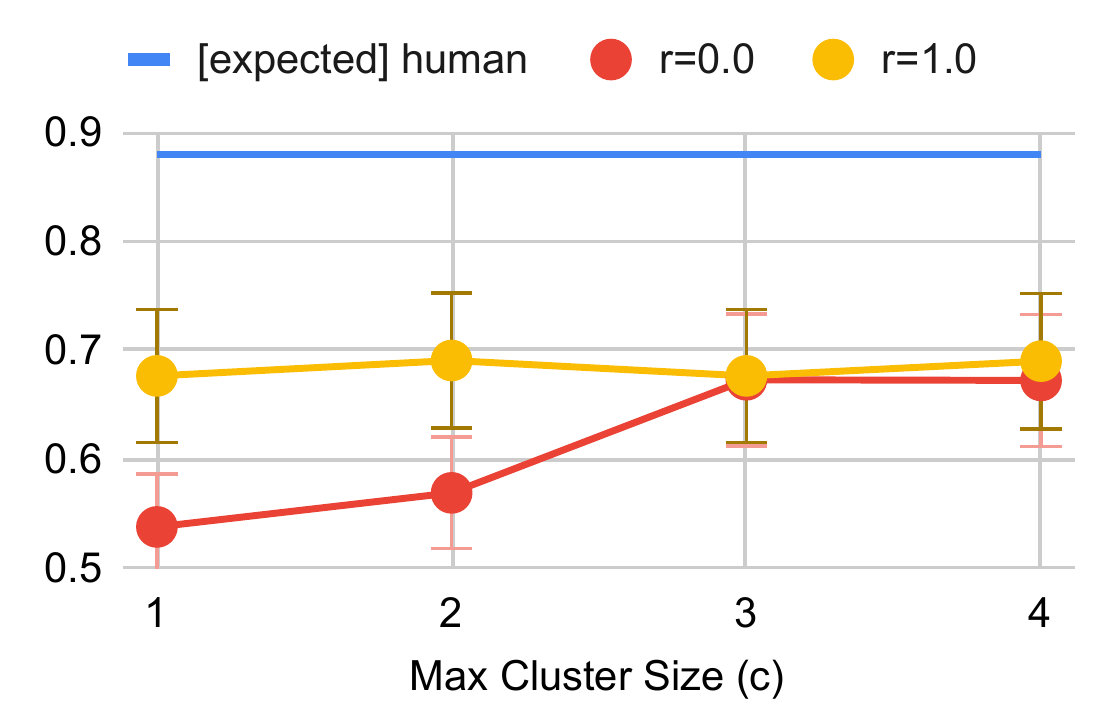}
    \end{subfigure}
    \begin{subfigure}[b]{0.32\textwidth}
        \centering
        \caption{\multirc}
        \includegraphics[scale=0.47,trim=0.2cm 0.5cm 0cm 0cm]{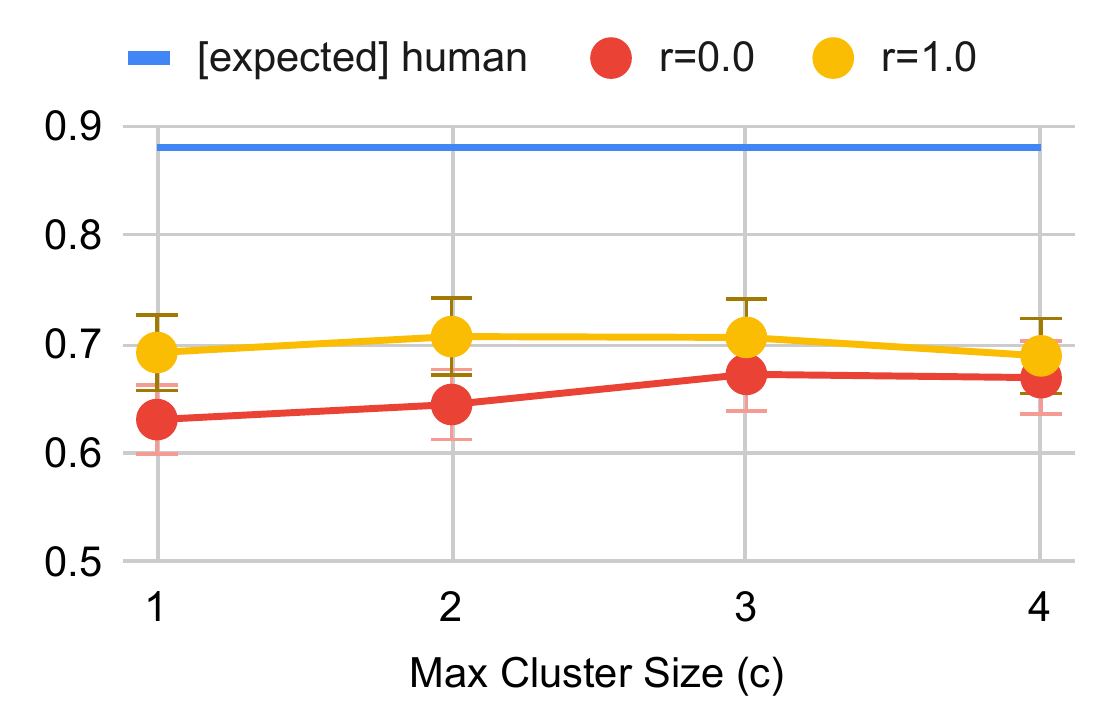}
    \end{subfigure}
    \begin{subfigure}[b]{0.32\textwidth}
        \centering
        \caption{\boolq}
        \includegraphics[scale=0.47,trim=0.2cm 0.5cm 0cm 0cm]{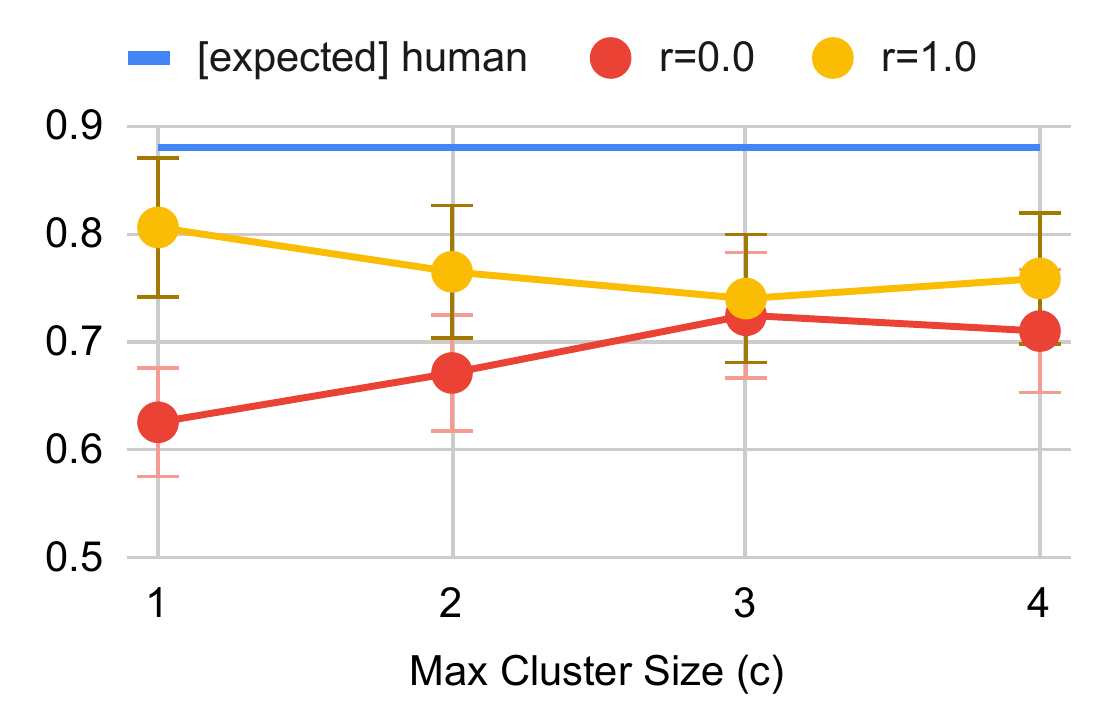}
    \end{subfigure}
    \caption{
        Model accuracy ($y$-axis) with a fixed budget $b$ and varying cluster size $c$ ($x$-axis), for two \daniel{extreme} cases: (i) \yellowtext{$r=1.0$} denoting a fixed total number of \emph{questions},
        (ii) \redtext{$r=0.0$} denoting a fixed total number of \emph{clusters}.
        \daniel{
            The plots indicate that including additional perturbations in each cluster \tushar{(going left to right)}, particularly when they are cheap \tushar{(closer to the $r=0$ case)}, adds value to the dataset by increasing model accuracy.
            }
        }
    \label{fig:cluster_eval}
\end{figure*}

\begin{figure*}[tb]
    \centering
    \begin{subfigure}[b]{0.32\textwidth}
        \centering
        \includegraphics[scale=0.47,trim=0.3cm 0.3cm 0cm 0cm]{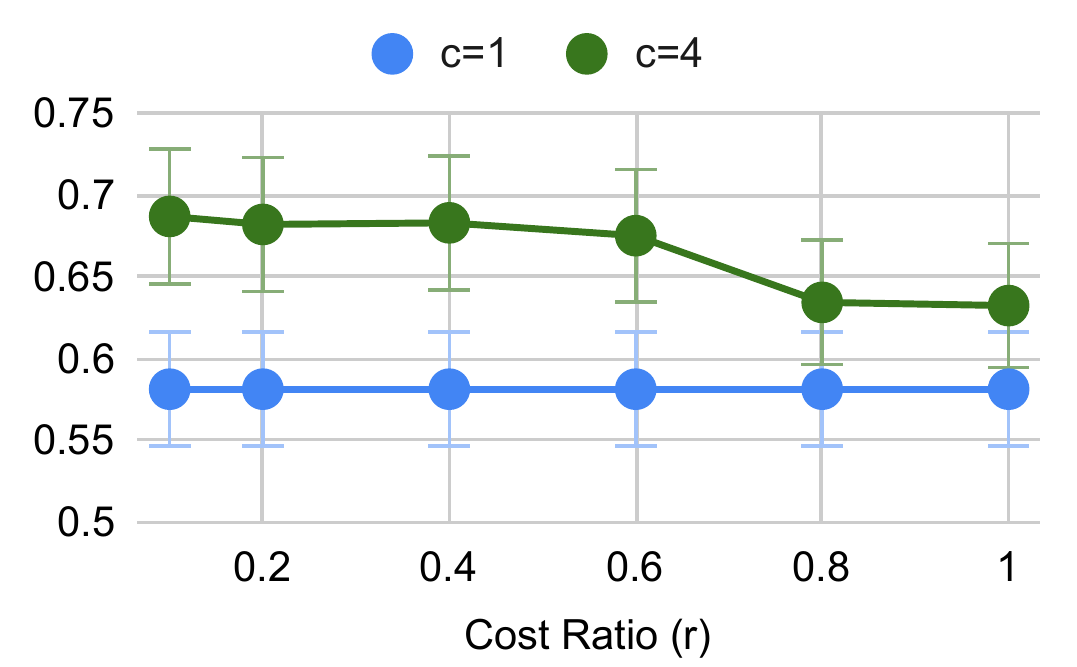}
    \end{subfigure}
    \begin{subfigure}[b]{0.32\textwidth}
        \centering
    \includegraphics[scale=0.47,trim=0.3cm 0.3cm 0cm 0cm]{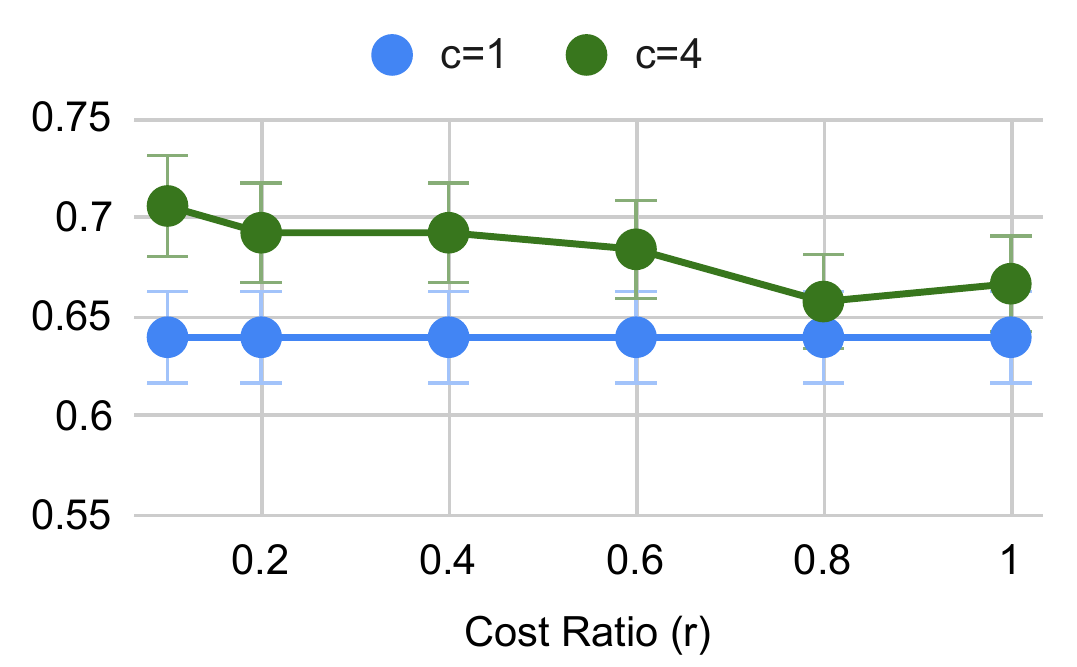}
    \end{subfigure}
    \begin{subfigure}[b]{0.32\textwidth}
        \centering
    \includegraphics[scale=0.47,trim=0.3cm 0.3cm 0cm 0cm]{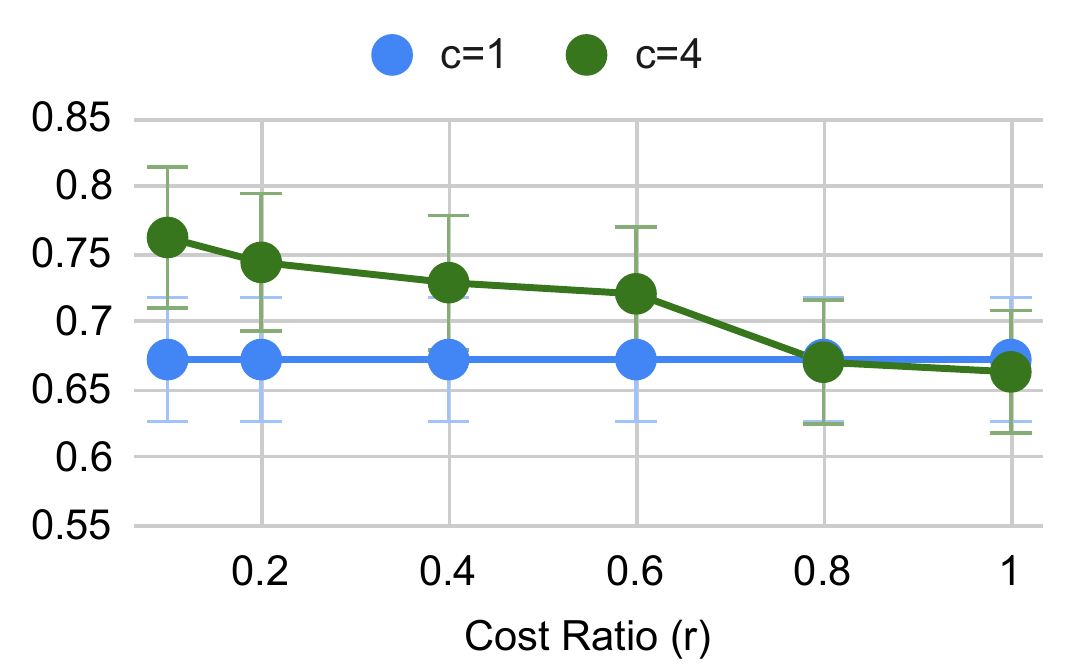}
    \end{subfigure}
    \caption{
        Model accuracy ($y$-axis) with a fixed total budget $b$ and varying cost ratio $r$ ($x$-axis), in two cases: (i) \bluetext{$c=1$} denoting singleton clusters \tushar{(standard approach)}, (ii) \greentext{$c=4$} 
        denoting cluster size 4 \tushar{(our approach)}.
        The smaller the cost, the higher the returns.
        \daniel{
            For a moderately low cost ratio such as 0.6, 
            the model trained on our perturbed datasets gains of
            3-5\% over a model trained on the traditionally constructed dataset. 
        }
    }
    \label{fig:cluster_value_with_cost}
\end{figure*}

\paragraph{Dataset subsampling.}
%
We sample questions from this expanded dataset to evaluate the value of perturbations as a function of different parameters. 
To simplify exposition, we will use the following notation. 
We assume a fixed budget $b$ for constructing the dataset where each new question costs $1$ unit, i.e., traditional methods would construct a dataset of size $b$ in the given budget. The perturbation cost ratio $r \leq 1$ is the cost of creating a perturbed question. When $r \approx 1$, perturbations are equally costly as writing out new instances. If $r \ll 1$, perturbations are cheap.
For instance, if $r=0.5$, each hand-written question costs the same as two perturbed questions. 

We denote the total number of instances and clusters with $N, C$, respectively. We use \boolqc{b,c,r} to denote the largest subset of \boolqR that can be generated with a total budget of $b$, with a \emph{maximum} cluster size of $c$, and relative cost ratio of $r$. In the special case where all clusters are of the exact same size $c$, these parameters are related as follows:
$$b =  \big(1 + (c-1)r\big) \times C,$$
where $1 + (c-1)r$ is the cost of a single cluster calculated as the cost of one seed examples and its $c-1$ perturbations.

To create \boolqc{b,c,r} we subsample  a maximum of $c$ questions from each perturbation cluster,  such that total number of clusters is no more than $\frac{b}{1 + (c-1)r}$ and the ratio of ``yes'' to ``no'' questions is 0.55.
Our subsampling starts with clusters of size at least $c$ and also considers smaller clusters if necessary.
\boolqc{b,1,r} (singleton clusters) corresponds to a dataset constructed in a similar fashion to \boolq, whereas \boolqc{b,4,r} (big clusters) roughly corresponds to the \boolqR\ dataset.

\section{Experiments}

To assess the impact of our perturbation approach, we evaluate 
standard RoBERTa-large model that has been shown to achieve state-of-the-art results on many tasks. 
Each experiment considers the effect of training on subsamples of \boolqR obtained under different conditions. 

Each of the points in the figures are averaged over 5 random subsampling of the dataset (with error bars to indicate the standard deviation). The Appendix includes further details about the setup as well as additional experiments.

We evaluate the QA model trained on various question sets on three test sets. (i) For assessing \textbf{robustness}, we use an expert-generated set \boolqExperts published in \citet{gardner2020contrastsets} with 339 high-quality perturbed questions based on \boolq. (ii) For assessing \textbf{generalization}, we use the subset of 260 training questions from \multirc~\cite{khashabi2018looking} that have binary (yes/no) answers, from training section of the their data.\footnote{The yes/no subset of dev was too small.} (c) The original \boolq test set, to ensure models trained on perturbed questions also \textbf{retain performance} on the original task.

\subsection{Effect of Cluster Size ($c$)}
\label{subsec:evaluation_wrt_cluster_size}

We study the value of clusters sizes in the perturbations in two \ashish{extreme} cases: (i) when perturbations cost the same as new questions ($r=1.0$) and the only limit is the our overall budget ($b=3.7k$), and (ii) when the perturbations cost negligible ($r=0.0$) but we are limited by the max cluster size $c$ and $b=1k$.\footnote{\ashish{In practice, we expect $r$ to lie somewhere in-between these two extremes, such as $r=0.3$ as discussed in \S\ref{subsec:evaluation_wrt_cost}.}}
For each case, we vary the max cluster size in the following rage: $[1, 2, 3, 4]$. As a result, in (i), $C$ vary from $3.7k$ to 951 ($N=3.7k$), and in (ii), $N$ vary from $1k$ to $4k$ ($C=1k$).


Fig.~\ref{fig:cluster_eval} shows the accuracy of models trained on these subsets across our three evaluation sets.
In \yellowtext{scenario (i)} with a fixed number of instances ($r=1$), it is evident that the size of the clusters (the number of perturbations) 
        does not
affect the model quality \daniel{(on 2 out of 3 datasets)}. 
This shows that perturbation clusters are equally informative as (traditional) independent instances. 
However, in \redtext{scenario (ii)} with a fixed number of clusters ($r=0$), the system performance consistently gets higher with larger clusters, even though the number of clusters is kept constant. This indicates that each additional perturbation adds value to the existing ones, especially in terms of model robustness and retaining performance on the original task. 

\subsection{Effect of Perturbations Cost Ratio ($r$)}
\label{subsec:evaluation_wrt_cost}

We now study the value of perturbations as a function of their cost ($r$). We vary this parameter within the range $(0, 1]$ for $b=1.5k$ and two max clusters sizes, $c=\{1, 4\}$. 
When \bluetext{$c=1$} (no perturbations), $N$ stays constant at $1.5k$. When \greentext{$c=4$}, $N$ varies from $4.6k$ to $1.5k$. Fig.~\ref{fig:cluster_value_with_cost} presents the accuracy of our model  as a function of $r$. 

While we don't know the exact crowdsourcing cost for \boolq, a typical question writing task might cost USD 0.60 per question. With our \ashish{perturbed dataset} costing USD 0.20 per question, we have r = 0.33. Given the same total budget $b = 1500$, we can thus infer from Fig.~\ref{fig:cluster_value_with_cost} that training on a dataset of perturbed questions would be about 10\% and 5\% more effective on \boolqExperts and \multirc, respectively.

The result on all datasets indicates that there is value in using perturbations clusters when $r \leq 0.6$, i.e., larger clusters can be more cost-effective for build better training sets. Even when they are not much cheaper, they retain the same performance as independent examples, making them a good alternative for dataset expansion given few sources of examples (e.g., low resource languages).

\section{Discussion}

\daniel{
A key question with respect to the premise of this work is whether the idea would generalize to other tasks. 
Here, we chose yes/no questions since this is the least-explored sub-area of QA (compared to extractive QA, for example) and hence could benefit from more efficient dataset construction. 
We (the authors) are cautiously optimistic that it would, although that is subject to factors such as the relative cost of creating diverse and challenging perturbations. Concurrent works have also explored a similar construction for other tasks but with different purposes~\cite{gardner2020contrastsets,kaushik2019learning}. 
}

\daniel{
We note that we assume a typical QA dataset construction process where workers write questions based on given fixed contexts~\cite{Rajpurkar2016SQuAD10}. This assumption may not always hold for alternative dataset generation pipelines, such as using an already available set of questions~\cite{kwiatkowski2019natural}. Even in such cases, one can still use the lessons learned here to apply \emph{natural perturbations} to a different stage in the annotation pipeline to make it more cost efficient. 
}

\section{Conclusion}

We proposed an alternative approach for constructing training sets, by expanding seed examples via natural perturbations. Our results demonstrate that models trained on perturbations of \boolq questions are more robust to minor variations and generalize better, while preserving performance on the original \boolq test set 
\daniel{as long as the natural perturbations are moderately cheap to create.}
 Creating perturbed examples is often cheaper than creating new ones and we empirically observed notable gains even at a moderate cost ratio of 0.6.

While this is \emph{not} a dataset paper (since our focus is on more on the value of natural perturbations for robust model design), we provide the natural perturbations resource for \boolq\ constructed during the course of this study.\footnote{https://github.com/allenai/natural-perturbations}

This work suggests a number of interesting lines of future investigation.
For instance, how do the results change as a function of the total dataset budget $b$ or large values of $c$?
Over-generation of perturbations can result in overly-similar (less-informative) variations of a seed example, making larger clusters valuable only up to a certain extent. While we leave a detailed study to future work, we expect general trends regarding the value of perturbations to hold broadly.

\subsection*{Acknowledgments}

The authors would like to thank Chandra Bhagavatula, Sihao Chen, and Niket Tandon for discussions and suggestions on early versions of this work, and anonymous Amazon Mechanical Turk workers for their help with annotation. 

\bibliography{cited_extra}
\bibliographystyle{acl_natbib}


\clearpage 

\appendix

\section{Question Perturbations: Further Details}
\label{sec:supp:perurbations_further}

We provide further details about the annotation. 

The task starts with a qualification step: we ask annotators to read a collection of meticulously designed instructions that describe the task. The annotators are allowed to participate, only after successfully passing the test included in the instructions. 

In addition, we restrict the task to ``Master'' workers from English-speaking countries (USA, UK, Canada, and Australia), at least 500 finished HITs and at least a 95\% acceptance rate.

Here is an screen cast of the relevant annotation interface interface: \url{https://youtu.be/MWbCRwanbOA}

During our earlier pilot experiments, we observed that the strategies used for perturbing ``yes'' questions tend to be different from those used for ``no'' questions. To make the task less demanding and help workers focus on a limited cognitive task, the annotation is done in two phases; one for ``yes'' questions, and another for  ``no'' questions.

Table~\ref{tab:statistics} provides a summary of \boolqR\ stats.

\input{tables/boolqr-stats}

\section{Details of \roberta Training}
\label{supp:sec:training_details}

We train the model on two-way questions using the input format: ``\texttt{[CLS]} passage \texttt{[SEP]} question \texttt{[SEP]} answer''. The model scores each answer (``yes'' or ``no'') by applying a linear classifier over the \texttt{[CLS]} representation for each answer's corresponding input. 
We train the linear classifier (and fine-tune \roberta\ weights) on the training sets and evaluate them on their corresponding dev/test sets.
We fixed the learning rate to 1e-5 as it generally performed the best on our datasets. 
We only varied the number of training epochs: \{7, 9, 11\} and effective batch sizes: {16, 32}. 
We chose this small hyper-parameter sweep to ensure that each model was fine-tuned using the same hyper-parameter sweep while not being prohibitively expensive. Each model was selected based on the best validation set accuracy. 
We report the numbers corresponding to the selected models on the test set. 

\section{Performances Across Datasets}
\label{sec:table_of_performances}

We compare a collection of solvers across our target datasets: 
the complete \boolqR dataset (dataset constructed from $\datasetSeed$ via perturbation), the original \boolq dataset, expert perturbations on \boolq, and binary subset of \multirc. 

The results are summarized in Table~\ref{tab:different:configurations}. 
Most of the rows are \roberta trained on a specified dataset. 
We have also included a row corresponding a system trained on the union of \boolq and \boolqR, referred to as \boolqpp for brevity. 
Most of the datasets are slightly skewed between the two classes, which is why the majority label baseline (\emph{Always-Yes} or \emph{Always-No}) achieves scores above 50\%. 
Rows indicated with * are reported directly from prior work. 
The human prediction on \boolqR is the majority label of 5 independent AMT annotators. The human performance on \boolq\ and \multirc\ are directly reported from SuperGLUE~\cite{wang2019superglue} leaderboard.\footnote{https://super.gluebenchmark.com/leaderboard/}

Here are the key observations in this table: 
\begin{itemize}
    \item While \roberta\ has almost human-level performance when trained and tested within \boolq, it suffers significant performance degradation when evaluated on other datasets (e.g., 68.7\% on \boolqR). 
    
    \item The systems fine-tuned on \boolqpp consistently generalize better across datasets. 
\end{itemize}

\begin{table}[ht]
    \centering
    \small
    \begin{tabular}{C{2cm}C{1.5cm}C{1.7cm}C{0.7cm}}
        \toprule
        Model &  Trained on & Evaluated on & Acc. \\
        \cmidrule(r){1-1} \cmidrule(r){2-2} \cmidrule(r){3-3} \cmidrule(r){4-4} 
            Human* & --- & \multirc & $\sim$83 \\ 
            \roberta & \boolqpp & \multirc & \textbf{78.8} \\ 
            \roberta & \boolqR & \multirc & 70.3 \\
            \roberta & \boolq & \multirc & 76.5 \\
            \emph{Maj-Vote} & --- & \multirc  & 63.4 \\
        \midrule
            Human* & --- & \boolq & 89.0 \\ 
            \roberta &  \boolqpp & \boolq & 85.5 \\ 
            \roberta &  \boolq & \boolq & \textbf{86.1} \\
            \roberta & \boolqR & \boolq & 78.6 \\
            \emph{Maj-Vote} & --- & \boolq  & 62.2 \\
        \midrule
            Human & --- & \boolqR & 89.4 \\
            \roberta & \boolqpp & \boolqR & \textbf{81.1} \\ 
            \roberta & \boolq & \boolqR & 68.7 \\
            \roberta & \boolqR & \boolqR & 78.4 \\
            \emph{Maj-Vote} & --- & \boolqR & 53.2 \\
        \midrule
            Human & --- & \boolqExperts & ? \\
            \roberta & \boolqpp & \boolqExperts & \textbf{76.4} \\ 
            \roberta & \boolq & \boolqExperts & 71.1 \\
            \roberta & \boolqR & \boolqExperts & 69.3 \\
            \emph{Maj-Vote} & --- & \boolqExperts & 50.7 \\
        \bottomrule
    \end{tabular}
    \caption{
    Various systems trained and evaluated on different datasets. Best non-human scores are in \textbf{bold}. Numbers in percentage.}
    \label{tab:different:configurations}
\end{table}




\section{Cluster-Level Evaluation}

An additional benefit of our approach is that it produces datasets with an inherent cluster structure. This enables the use of metrics such as ConsensusScore~\cite{Shah2019CycleConsistencyFR} to evaluate the extent to which a model acts \emph{consistently} within each cluster, which provides another measure of robustness.

While evaluation measures are often computed on \emph{per-instance} level, the cluster structure of  \boolqR enables us to provide \emph{per-cluster} metrics of quality. In particular, we are interested in the following question: \emph{to what extent do our models act consistently across questions in each cluster?}

To measure this, we use the \emph{consensus score} $CS(k)$ introduced by \namecite{Shah2019CycleConsistencyFR}. 
For an integer parameter $k \geq 1$, the score $CS(k)$ for a single cluster $C$ is defined as the fraction of size-$k$ sub-clusters of $C$
where the model answers \emph{all} instances correctly. The $CS(k)$ score for a clustered dataset is the average of these scores across all clusters. Intuitively, $k=1$ represents the traditional un-clustered accuracy (assuming all clusters with the same size). As $k$ grows to reach the cluster size, models must answer the \emph{entire} cluster correctly in order to score positively on that cluster.

\begin{figure}
    \centering
    \includegraphics[scale=0.43,trim=0.8cm 1.0cm 0.5cm 0.5cm]{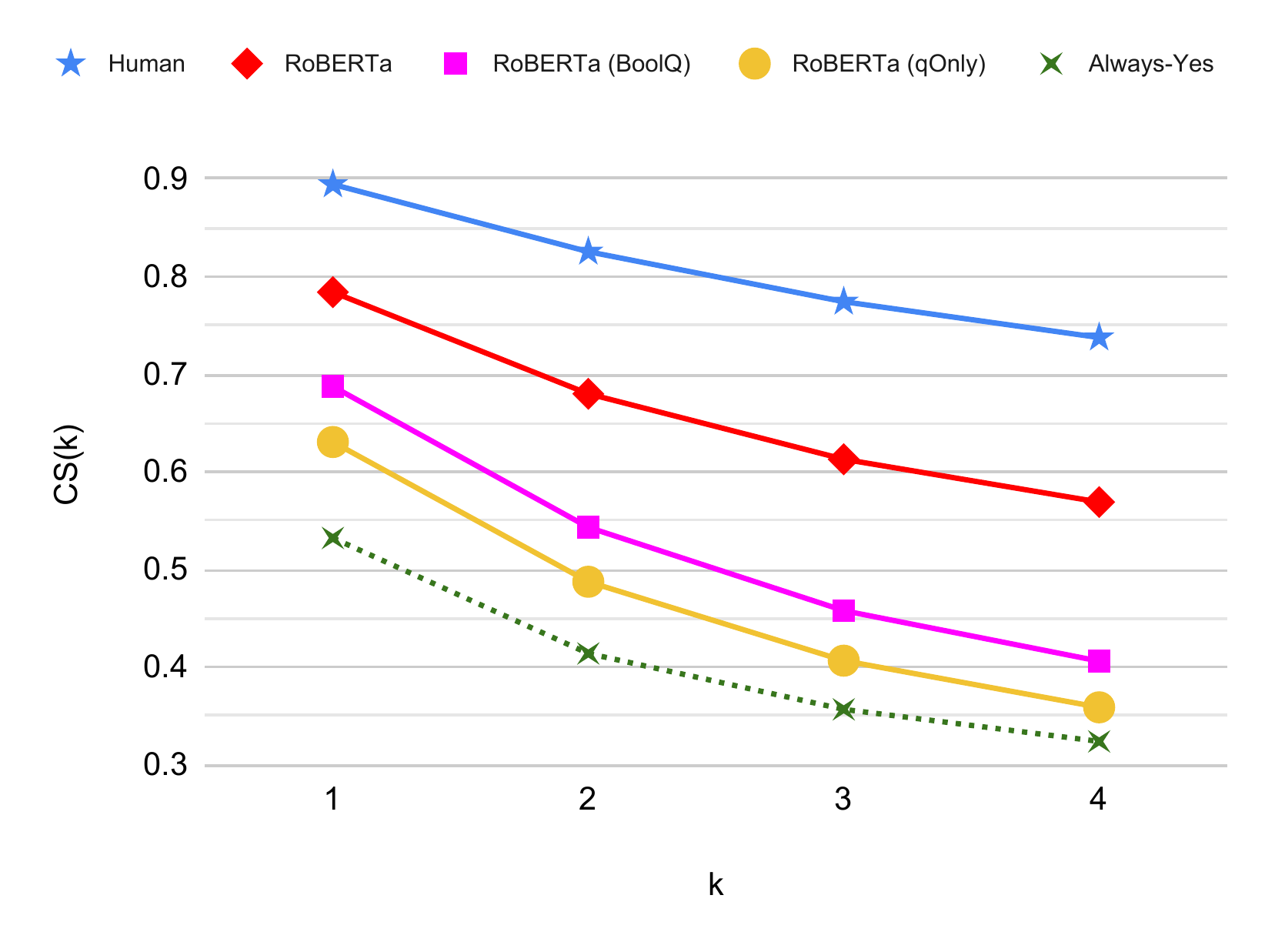}
    \caption{Consensus metric $CS(k)$ on the y-axis for various values of $k$ on the x-axis.}
    \label{fig:k:robustness}
\end{figure}

We plot this score for $k \in \{1, 2, 3, 4\}$ for various QA models in Fig~\ref{fig:k:robustness}. While all the models (including human) have deceasing consensus score for larger values of $k$, machine solvers have a steeper slope compared to human. As a result, we have an even larger gap of 17\% between human-\roberta (at $k=4$), when 
evaluated on their consistency.

\section{Rule-Based Perturbations}
\label{supp:manualperturbations}

An alternate way to get cheap perturbations would be to use rule-based paraphrase systems --- which are arguably cheaper than human-annotated perturbations. 

Our intuition is that rule-based perturbations generally have simplistic definitions and hence, rarely benefit general reasoning problems in language.  
Interesting and diverse rule-based perturbations can be difficult to develop, and existing approaches are often reverse-engineered by QA models. Further, unlike our proposal, automatic perturbation approaches, such as question rephrasing, generally preserve the answer and do not use the provided context the question is referring to, limiting their richness.

That being said, we put some effort into developing rule-based/machine-generated baselines for comparison. However, since these efforts did not result in any reasonably sophisticated baselines, we decided to not include them in the main text. 

Here we're showing examples of perturbations generated via a recent machine paraphraser system.\footnote{https://github.com/decompositional-semantics-initiative/improved-ParaBank-rewriter }

\begin{mdframed}
\small 
\textbf{Original Question:}
Will there be a season 4 of da vinci’s demons? FALSE
\end{mdframed}

The corresponding machine-perturbed questions are: 

\begin{mdframed}
\small \noindent
Will there be a season four of da vinci demons? FALSE \\ 
Will there be season four of da vinci demons? FALSE \\ 
Is there a season four of da vinci demons? FALSE\\ 
Is there gonna be a season four of da vinci demons? FALSE
\end{mdframed}

These automated perturbations stand in contrast with our human-perturbed questions, which also take the provided context into account:

\begin{mdframed}
\small \noindent
Was there a season 3 of da vinci's demons? TRUE \\ 
There be a season 4 of da vinci's demons? FALSE \\ 
Will there be no season 4 of da vinci's demons? TRUE 
\end{mdframed}

As evident by the example, the machine-generated perturbations are generally minor and, not surprisingly, did not provide a useful enough signal to the model to improve its accuracy. We are open to suggestions if the reviewers have any suggestion on creating more reasonable rule-based perturbation baselines.

\end{document}

%% file: tables/boolqr-stats.tex
\begin{table}[ht]
    \small
    \centering
    \footnotesize
    \begin{tabular}{lcrrrr}
        \toprule
        \multicolumn{2}{l}{Measure}                     & Full & Train & Dev & Test \\
        \cmidrule(r){1-3}  \cmidrule(r){4-4} \cmidrule(r){5-5} \cmidrule(r){6-6}
        \multicolumn{2}{l}{\# of questions}             & 17,323 & 9727 &  4434 & 3162 \\
        \multicolumn{2}{l}{\# of ``yes'' questions}     & 9,724 & 5733 & 2263 & 1728 \\
        \multicolumn{2}{l}{\# of ``no'' questions}      & 7,599 & 3994 & 2171 & 1434 \\
        \midrule
        \multicolumn{2}{l}{\# of clusters}              & 4064 & 2408 & 919 & 737 \\
        \multicolumn{2}{l}{average cluster size}        & 4.3 & 4.1 & 4.8 & 4.3 \\
        \multicolumn{2}{l}{median cluster size}         & 3.0 & 3.0 & 3.0 & 3.0 \\
        \bottomrule
    \end{tabular}
    \caption{
        Statistics of \boolqR. 
    }
    \label{tab:statistics}
\end{table}

%% file: main.bbl
\providecommand{\CNFX}[1]{
  {\em{\textrm{(#1)}}}}\providecommand{\CNFSoCG}{\CNFX{SoCG}}\providecommand{\CNFCCCG}{\CNFX{CCCG}}\providecommand{\CNFFOCS}{\CNFX{FOCS}}\providecommand{\CNFSODA}{\CNFX{SODA}}\providecommand{\CNFSTOC}{\CNFX{STOC}}\providecommand{\CNFPODS}{\CNFX{PODS}}\providecommand{\CNFISAAC}{\CNFX{ISAAC}}\providecommand{\CNFFSTTCS}{\CNFX{FSTTCS}}\providecommand{\CNFIJCAI}{\CNFX{IJCAI}}\providecommand{\CNFBROADNETS}{\CNFX{BROADNETS}}\providecommand{\CNFCCC}{\CNFX{CCC}}\providecommand{\CNFECCC}{\CNFX{ECCC}}\providecommand{\CNFEC}{\CNFX{EC}}\providecommand{\CNFITCS}{\CNFX{ITCS}}\providecommand{\CNFIPCO}{\CNFX{IPCO}}\providecommand{\CNFKDD}{\CNFX{KDD}}\providecommand{\CNFLICS}{\CNFX{LICS}}\providecommand{\CNFEC}{\CNFX{EC}}\providecommand{\CNFICALP}{\CNFX{ICALP}}\providecommand{\CNFESA}{\CNFX{ESA}}
\begin{thebibliography}{21}
\expandafter\ifx\csname natexlab\endcsname\relax\def\natexlab#1{#1}\fi

\bibitem[{Bowman et~al.(2015)Bowman, Angeli, Potts, and
  Manning}]{bowman2015snli}
S.~Bowman, G.~Angeli, C.~Potts, and C.~Manning. 2015.
\newblock A large annotated corpus for learning natural language inference.
\newblock In \emph{Procedings of EMNLP}.

\bibitem[{Cheng et~al.(2018)Cheng, Tu, Meng, Zhai, and Liu}]{cheng2018towards}
Y.~Cheng, Z.~Tu, F.~Meng, J.~Zhai, and Y.~Liu. 2018.
\newblock Towards robust neural machine translation.
\newblock In \emph{Proceedings of ACL}.

\bibitem[{Clark et~al.(2019)Clark, Lee, Chang, Kwiatkowski, Collins, and
  Toutanova}]{clark2019boolq}
C.~Clark, K.~Lee, M-W. Chang, T.~Kwiatkowski, M.~Collins, and K.~Toutanova.
  2019.
\newblock {BoolQ}: {E}xploring the surprising difficulty of natural yes/no
  questions.
\newblock In \emph{Proceedings of NAACL}.

\bibitem[{Dinan et~al.(2019)Dinan, Humeau, Chintagunta, and
  Weston}]{dinan2019build}
E.~Dinan, S.~Humeau, B.~Chintagunta, and J.~Weston. 2019.
\newblock Build it break it fix it for dialogue safety: Robustness from
  adversarial human attack.
\newblock In \emph{Proceedings of EMNLP-IJCNLP}, pages 4529--4538.

\bibitem[{Ebrahimi et~al.(2018)Ebrahimi, Lowd, and
  Dou}]{ebrahimi2018adversarial}
J.~Ebrahimi, D.~Lowd, and D.~Dou. 2018.
\newblock On adversarial examples for character-level neural machine
  translation.
\newblock In \emph{Proceedings of COLING}.

\bibitem[{Gardner et~al.(2020)}]{gardner2020contrastsets}
M.~Gardner et~al. 2020.
\newblock Evaluating models’ local decision boundaries via contrast sets.
\newblock In \emph{Proceedings of EMNLP}.

\bibitem[{Geva et~al.(2019)Geva, Goldberg, and Berant}]{geva2019we}
M.~Geva, Y.~Goldberg, and J.~Berant. 2019.
\newblock Are we modeling the task or the annotator? {A}n investigation of
  annotator bias in natural language understanding datasets.
\newblock In \emph{Proceedings of EMNLP-IJCNLP}.

\bibitem[{Glockner et~al.(2018)Glockner, Shwartz, and
  Goldberg}]{glockner2018breaking}
M.~Glockner, V.~Shwartz, and Y.~Goldberg. 2018.
\newblock Breaking nli systems with sentences that require simple lexical
  inferences.
\newblock In \emph{Proceedings of ACL}.

\bibitem[{Hu et~al.(2019)Hu, Khayrallah, Culkin, Xia, Chen, Post, and
  Van~Durme}]{rewriting}
J.~Edward Hu, Huda Khayrallah, Ryan Culkin, Patrick Xia, Tongfei Chen, Matt
  Post, and Benjamin Van~Durme. 2019.
\newblock Improved lexically constrained decoding for translation and
  monolingual rewriting.
\newblock In \emph{Proceedings of NAACL-HLT}.

\bibitem[{Jia and Liang(2017)}]{Jia2017AdversarialEF}
R.~Jia and P.~Liang. 2017.
\newblock Adversarial examples for evaluating reading comprehension systems.
\newblock In \emph{Proceedings of {EMNLP}}.

\bibitem[{Kaushik et~al.(2020)Kaushik, Hovy, and Lipton}]{kaushik2019learning}
D.~Kaushik, E.~Hovy, and Z.~C. Lipton. 2020.
\newblock Learning the difference that makes a difference with
  counterfactually-augmented data.
\newblock In \emph{Proceedings of ICLR}.

\bibitem[{Khashabi et~al.(2018)Khashabi, Chaturvedi, Roth, Upadhyay, and
  Roth}]{khashabi2018looking}
D.~Khashabi, S.~Chaturvedi, M.~Roth, S.~Upadhyay, and D.~Roth. 2018.
\newblock Looking beyond the surface: A challenge set for reading comprehension
  over multiple sentences.
\newblock In \emph{Proceedings of NAACL}.

\bibitem[{Khashabi et~al.(2016)Khashabi, Khot, Sabharwal, Clark, Etzioni, and
  Roth}]{tableilp2016:ijcai}
D.~Khashabi, T.~Khot, A.~Sabharwal, P.~Clark, O.~Etzioni, and D.~Roth. 2016.
\newblock Question answering via integer programming over semi-structured
  knowledge.
\newblock In \emph{Proceedings of {IJCAI}}.

\bibitem[{Kobayashi(2018)}]{kobayashi2018contextual}
S.~Kobayashi. 2018.
\newblock Contextual augmentation: Data augmentation by words with paradigmatic
  relations.
\newblock In \emph{Proceedings of NAACL}.

\bibitem[{Kwiatkowski et~al.(2019)Kwiatkowski, Palomaki, Redfield, Collins,
  Parikh, Alberti, Epstein, Polosukhin, Devlin, Lee
  et~al.}]{kwiatkowski2019natural}
T.~Kwiatkowski, J.~Palomaki, O.~Redfield, M.~Collins, A.~Parikh, C.~Alberti,
  D.~Epstein, I.~Polosukhin, J.~Devlin, K.~Lee, et~al. 2019.
\newblock Natural questions: a benchmark for question answering research.
\newblock \emph{TACL}, 7:453--466.

\bibitem[{Levesque et~al.(2011)Levesque, Davis, and
  Morgenstern}]{levesque2011winograd}
H.~J. Levesque, E.~Davis, and L.~Morgenstern. 2011.
\newblock The {Winograd} schema challenge.
\newblock In \emph{AAAI Spring Symposium: Logical Formalizations of Commonsense
  Reasoning}.

\bibitem[{Peng et~al.(2015)Peng, Khashabi, and Roth}]{PengKhRo15}
H.~Peng, D.~Khashabi, and D.~Roth. 2015.
\newblock Solving hard coreference problems.
\newblock In \emph{Proceedings of NAACL}.

\bibitem[{Rajpurkar et~al.(2016)Rajpurkar, Zhang, Lopyrev, and
  Liang}]{Rajpurkar2016SQuAD10}
P.~Rajpurkar, J.~Zhang, K.~Lopyrev, and P.~Liang. 2016.
\newblock Squad: 100, 000+ questions for machine comprehension of text.
\newblock In \emph{Proceedings of EMNLP}.

\bibitem[{Sakaguchi et~al.(2020)Sakaguchi, Bras, Bhagavatula, and
  Choi}]{sakaguchi2019winogrande}
K.~Sakaguchi, R.~L. Bras, C.~Bhagavatula, and Y.~Choi. 2020.
\newblock {WINOGRANDE}: {A}n adversarial {W}inograd schema challenge at scale.
\newblock In \emph{Proceedings of AAAI}.

\bibitem[{Shah et~al.(2019)Shah, Chen, Rohrbach, and
  Parikh}]{Shah2019CycleConsistencyFR}
M.~Shah, X.~Chen, M.~Rohrbach, and D.~Parikh. 2019.
\newblock Cycle-consistency for robust visual question answering.
\newblock In \emph{Proceedings of CVPR}.

\bibitem[{Wang et~al.(2019)Wang, Pruksachatkun, Nangia, Singh, Michael, Hill,
  Levy, and Bowman}]{wang2019superglue}
A.~Wang, Y.~Pruksachatkun, N.~Nangia, A.~Singh, J.~Michael, F.~Hill, O.~Levy,
  and S.~Bowman. 2019.
\newblock Superglue: A stickier benchmark for general-purpose language
  understanding systems.
\newblock In \emph{Proceedings of NourIPS}.

\end{thebibliography}
